\documentclass[11pt,a4paper]{article}
\usepackage{times,latexsym}
\usepackage{url}
\usepackage[T1]{fontenc}

\usepackage[acceptedWithA]{tacl2021v1}

\usepackage[utf8]{inputenc}

\usepackage{graphicx}

\usepackage{times}
\usepackage{latexsym}
\usepackage{microtype}
\usepackage{graphicx}
\usepackage{adjustbox}
\usepackage{multirow}
\usepackage[normalem]{ulem}
\usepackage{xcolor,colortbl}
\usepackage{soul}
\setuldepth{Berlin}
\usepackage{paralist}
\usepackage{xfrac}
\usepackage[inline]{enumitem}
\usepackage{graphicx}
\usepackage{url}
\usepackage{textcomp}
\usepackage{caption}
\usepackage{subcaption}
\usepackage{color,soul}
\usepackage{paralist}
\usepackage{float}
\usepackage{placeins}
\usepackage{aliascnt}
\usepackage{mathtools}
\usepackage{amsmath} 
\usepackage{hhline}
\usepackage{booktabs}
\usepackage{footnote}
\usepackage{soul}
\usepackage{bbm}
\usepackage{amssymb}
\usepackage{pifont}
\usepackage{adjustbox}
\usepackage{array}
\usepackage{lipsum}
\usepackage{scrextend}

\newcommand\setalign{4pt}
\newcolumntype{R}[2]{%
    >{\adjustbox{angle=#1,lap=\width-(#2)}\bgroup}%
    l%
    <{\egroup}%
}

\usepackage{xspace,mfirstuc,tabulary}

\newif\iftaclinstructions
\taclinstructionsfalse 
\iftaclinstructions
\renewcommand{\confidential}{}
\renewcommand{\anonsubtext}{(No author info supplied here, for consistency with
TACL-submission anonymization requirements)}
\newcommand{\instr}
\fi

\iftaclpubformat 

\else

\fi

\definecolor{colorsmall}{rgb}{1,1,1}
\definecolor{colormedium}{rgb}{1,1,1}
\definecolor{colorlarge}{rgb}{1,1,1}
\definecolor{colorgt}{rgb}{0.9, 0.9, 0.9}

\newcommand{\smallmodel}{\cellcolor{colorsmall}}
\newcommand{\mediummodel}{\cellcolor{colormedium}}
\newcommand{\largemodel}{\cellcolor{colorlarge}}

\newcommand\blfootnote[1]{%
  \begingroup
  \renewcommand\thefootnote{}\footnote{#1}%
  \addtocounter{footnote}{-1}%
  \endgroup
}

\title{An Efficient Self-Supervised Cross-View Training For Sentence Embedding}

\author{
Peerat Limkonchotiwat\textsuperscript{\dag},  Wuttikorn Ponwitayarat\textsuperscript{\dag}, Lalita Lowphansirikul\textsuperscript{\dag},\\
\textbf{Can Udomcharoenchaikit}\textsuperscript{\dag}, \textbf{Ekapol Chuangsuwanich}\textsuperscript{‡}, \textbf{Sarana Nutanong}\textsuperscript{\dag *}\\
  \textsuperscript{\dag}School of Information Science and Technology, VISTEC, Thailand\\
  \textsuperscript{‡}Department of Computer Engineering,  Chulalongkorn University, Thailand \\
  \texttt{\{peerat.l\_s19,wuttikorn.p\_s22,lalita.l\_s22}
  \\
  \texttt{,canu\_pro,snutanon\}@vistec.ac.th,}\\
    \texttt{ekapolc@cp.eng.chula.ac.th}
  }

\date{}

\begin{document}

\maketitle
\blfootnote{\textsuperscript{*}Corresponding Author}

\begin{abstract}
  Self-supervised sentence representation learning is the task of constructing an embedding space for sentences without relying on human annotation efforts.
One straightforward approach is to finetune a pretrained
language model (PLM) with a representation learning method such as \emph{contrastive learning}.
While this approach achieves impressive performance on larger PLMs, the performance rapidly degrades as the number of parameters decreases.
In this paper, we propose a framework called \emph{Self-supervised Cross-View Training (SCT)} to narrow the performance gap between large and small PLMs. 
To evaluate the effectiveness of SCT, we compare it to 5 baseline and state-of-the-art competitors on seven \emph{Semantic Textual Similarity (STS)} benchmarks using 5 PLMs with the number of parameters ranging from 4M to 340M.
The experimental results show that STC outperforms the competitors for PLMs with less than 100M parameters in 18 of 21 cases.\footnote{Codes and Models: \url{https://github.com/mrpeerat/SCT}}

\end{abstract}

\section{Introduction}
Self-supervised sentence representation learning is the task of constructing an embedding space for sentences without relying on human annotation efforts. 
Recent advancements in self-supervised sentence representation present promising results on various downstream tasks such as Semantic Textual Similarity (STS) and text classification.
For example, \citet{gao-etal-2021-simcse} found that self-supervised sentence embedding methods could be on par with supervised methods~\cite{reimers-gurevych-2019-sentence} on various STS benchmarks.

A straightforward approach to self-supervised sentence representation is to finetune a pre-trained language model (PLM), i.e., BERT~\cite{DBLP:conf/naacl/DevlinCLT19} and RoBERTa~\cite{Liu-etal-2019-roberta}, with a representation learning technique. 
One popular method is contrastive learning.
This learning method enables self-supervised representation learning by creating a self-referencing mechanism through data augmentation~\cite{gao-etal-2021-simcse,zhang-etal-2022-virtual,zhou-etal-2022-debiased,klein-nabi-2022-scd, yan-etal-2021-consert,liu-etal-2021-fast, kim-etal-2021-self,cao-etal-2022-exploring}.
These works have demonstrated improvements over existing self-supervised techniques in sentence embedding benchmark datasets (i.e., STS and text classification). 

\begin{figure*}[ht]
\hspace*{-2mm}
\centering
\includegraphics[width=1.13\textwidth]{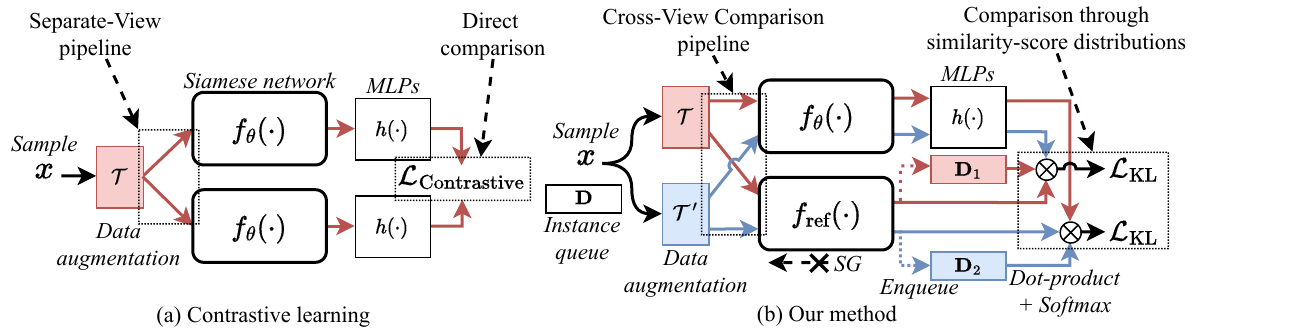}
\vspace{-6mm}
\caption{(a) The overview of self-supervised contrastive learning for sentence embedding. Contrastive learning is applied to directly compare the input $x$ produced from the separate-view pipeline $\mathcal{T}$.
(b) The \emph{Self-Supervised Cross-View Training (SCT)} pipeline. 
We calculate similarity score distributions between two networks ($f_\theta$ and $f_\text{ref}$) from the cross-view pipeline and minimize them through similarity-score distribution. In addition, the two networks do not require identical architecture nor share weights. They can be large and small networks (distillation) or Siamese networks.}
\vspace{-4mm}
\label{fig:simple_overall}
\end{figure*}

\begin{figure}[htbp]
\vspace{-3mm}
\centering
\includegraphics[width=0.49\textwidth]{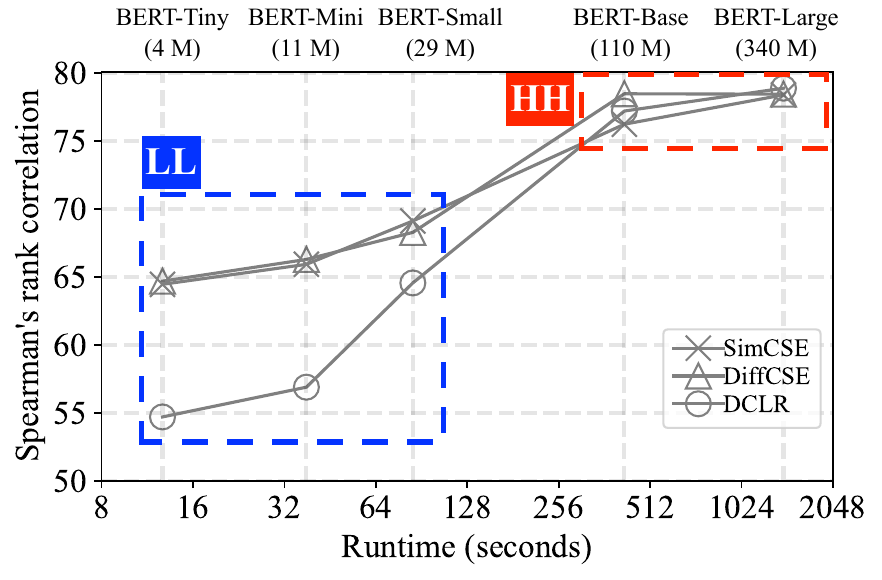}
\vspace{-8mm}
\caption{\label{font-table} Comparison between sentence representation methods on different model sizes. We averaged Spearman's rank correlation across seven STS datasets. LL denotes the low-cost, low-performance group, and HH denotes the high-cost, high-performance group.}
\vspace{-2mm}
\label{fig:overall_score}
\end{figure}

Figure~\ref{fig:overall_score} shows how three existing methods SimCSE~\cite{gao-etal-2021-simcse}, DiffCSE~\cite{chuang-etal-2022-diffcse}, and DCLR~\cite{zhou-etal-2022-debiased} perform on the BERT architecture as we varied the number of parameters from 4M to 340M.
While these self-supervised techniques achieve impressive performance on larger PLMs (i.e., those with more than 100M parameters), the performance rapidly degrades as the number of parameters decreases~\cite{wu2021disco,zhang-etal-2022-virtual,limkonchotiwat-etal-2022-congen}.
The figure also shows how the data points organize themselves into two distinct groups: LL and HH.
%
%
%
%
\begin{compactitem}[\hspace{\setalign}•]
\item \emph{High Cost, High Performance (HH).} As shown in Figure~\ref{fig:overall_score}, all models in this group, i.e., BERT-Base and BERT-Large, score more than 75, with the inference times over 420.9 seconds regardless of the learning method. 
%
%

\item \emph{Low Cost, Low Performance (LL).} This group contains all methods on models with less than 100 parameters, i.e., BERT-Tiny, BERT-Mini, and BERT-Small. 
All models in this group score less than 70, with the inference times less than 84.7 seconds regardless of the learning method.

\end{compactitem}
Despite the apparent benefit of low computation costs, smaller models, i.e., BERT-Tiny, BERT-Mini, and BERT-Small, are often neglected.  
Greater emphasis should be placed on exploring the potential to enhance the performance of smaller models through novel learning methods specifically tailored to their unique characteristics.

In this paper, we propose a framework called \emph{Self-Supervised Cross-View Training (SCT)} to narrow the performance gaps between large and small PLMs. 
Figure~\ref{fig:simple_overall} displays the difference between the traditional contrastive learning approach and ours.
As shown in Figure~\ref{fig:simple_overall}a, the two views are separated for contrastive learning, and the outputs from $h(\cdot)$ are directly compared to each other.
Figure~\ref{fig:simple_overall}b highlights the key distinctions of SCT based on two concepts: \emph{cross-view comparison} and \emph{similarity-score-distribution learning}. 
\begin{compactitem}[\hspace{\setalign}•]
    
\item \emph{Cross-view comparison}: The ability to self-reference is crucial to self-supervised learning. We derive a novel mechanism for two different augmented views to reference each other.  

\item \emph{Similarity-score-distribution learning}: The way we quantify loss is critical to any learning process. Our method calculates the loss by measuring the discrepancy between two similarity score distributions obtained from cross-comparing two different views. 
\end{compactitem}
The combination of these two concepts provides additional guidance which improves the effectiveness of self-supervised sentence representation learning on small PLMs.

To evaluate the effectiveness of SCT, we compare it to state-of-the-art (SOTA) competitors on STS, re-ranking, and natural language inference (NLI) benchmarks.
We also employ a distillation setting using BERT-Large-SimCSE as a teacher model.
The experimental results on STS demonstrated that our framework could address the drastic performance degradation problems in small PLMs by outperforming competitors in every case when the number of parameters is less than 100M.
For the smallest model (\#parameter: 4 million), we improved the performance from 64.47 to 69.73 points compared to SimCSE.
In the case of large PLMs (i.e., those with more than 100M parameters), our model's performance was on par with the current SOTA model when tested on BERT-Base and BERT-Large.
For the distillation setting, we outperformed all distillation competitors on all PLMs. 
For the re-ranking and NLI tasks, we improved the downstream tasks' performance for nearly all settings.

The contributions of our work are as follows:
\begin{compactitem}[\hspace{\setalign}•]
    \item We formulate a cross-view comparison pipeline to provide a more robust self-referencing mechanism for self-supervised sentence representation learning on smaller PLMs (those with less than 100M parameters). 
    
    \item Based on the cross-view comparison, we propose a method to measure the discrepancies between the cross-view outputs by comparing their respective similarity score distributions rather than the direct outputs. 
    
    \item We evaluate the effectiveness of SCT against five competitors on three families of PLMs using STS and downstream benchmark datasets.
    In addition, we also provide an in-depth analysis of different components in the cross-view pipeline to assess their effectiveness individually. 

\end{compactitem}

\section{Related Work}
\label{section_related_work}
Self-supervised learning is becoming more popular as a method to learn sentence representation from pre-trained language models (PLMs) without annotated information from training corpora.
We cover well-known self-supervised sentence representation learning techniques in the following subsections.

\subsection{Contrastive Learning} \label{subsec:contrast}
Contrastive learning constructs an embedding space by treating augmentations of an anchor as positives and other samples as negatives. 
The anisotropic problem is addressed by pulling a positive sample and pushing a negative sample with respect to an anchor sample.
\citet{gao-etal-2021-simcse} showed that the way we obtain positive and negative samples is critical to the performance of the representation.
\citet{kim-etal-2021-self,cao-etal-2022-exploring} utilized a different PLM to generate positive and negative samples for each anchor. 
\citet{fang2020cert} derived a method using two back-translations to create two different augmented views.
Another popular approach is to generate positive and negative pairs using feature dimension dropouts~\cite{gao-etal-2021-simcse,yan-etal-2021-consert,liu-etal-2021-fast,klein-nabi-2022-scd}. 
The experimental results from these works outperformed the traditional self-supervised sentence embedding methods.

A more advanced technique uses an additional function to help distinguish positive from negative samples. 
For example, \citet{zhou-etal-2022-debiased} proposed an additional debias function by mapping negative samples to the Gaussian distribution while individually assigning a weight to each contrastive negative sample. 
\citet{zhang-etal-2022-virtual} proposed a virtual augmentation scheme by approximating the nearest neighborhood from the neighboring samples to create the virtual negative samples.
\citet{chuang-etal-2022-diffcse} introduced a discriminator network to contrastive learning by classifying whether each word in a sentence is edited. 
Although these works demonstrated good performance, contrastive learning requires a judicial consideration of negative sampling to prevent false negatives.

\subsection{Learning Without Negative Samples} \label{subsec:without_neg}
A popular method to avoid false negatives is to design a learning process that uses only positive samples.  
BSL~\cite{zhang-etal-2021-bootstrapped} adapted BYOL's learning algorithm~\cite{gril-etal-2020-byol}, which maximizes the similarity between two augmented views of each sentence. In particular, BSL created two augmented views from a PLM.
The method uses a weighted exponential moving average of embeddings as a self-referencing mechanism. 
\citet{klein-nabi-2022-scd} adapted a redundancy representation learning algorithm from \citet{pmlr-v139-zbontar21a} and added a cosine similarity to maximize the similarity between the two samples formulated from high and low intense feature-dropout rate models.
While these methods allow us to perform self-supervised learning without negative samples, they are still outperformed by contrastive learning. 

\subsection{Sentence Representation Distillation}\label{subsec:sent_distil}
Distillation is a widely used technique for creating a small PLM (student) from an existing large PLM (teacher)~\cite{turc2019,DBLP:conf/nips/WangW0B0020}. 
Several sentence representation works proposed self-supervised distillation frameworks.
For instance, \citet{wu2021disco} proposed an self-supervised contrastive distillation. 
They formulated an anchor and other components (positives and negatives) of contrastive learning using a small and large PLMs, respectively. 
\citet{limkonchotiwat-etal-2022-congen} proposed a distillation framework based on the instance queue concept. 
A large PLM formulated representations for an instance queue, while the small PLM mimicked the relation between its representations and those in the instance queue.

These methods have been shown to reduce the performance gap between a small and large PLMs effectively.
However, none of the sentence representation works present how to decrease the gap without utilizing a large PLM.
This research question is an important problem that needs to be addressed, especially since utilizing a large PLM may not always be feasible in practice. 
Therefore, it is crucial to propose techniques that can decrease the gap with or without utilizing knowledge from a large PLM.

\begin{figure*}[t]
\hspace*{-2mm}
\centering
\includegraphics[width=1.13\textwidth]{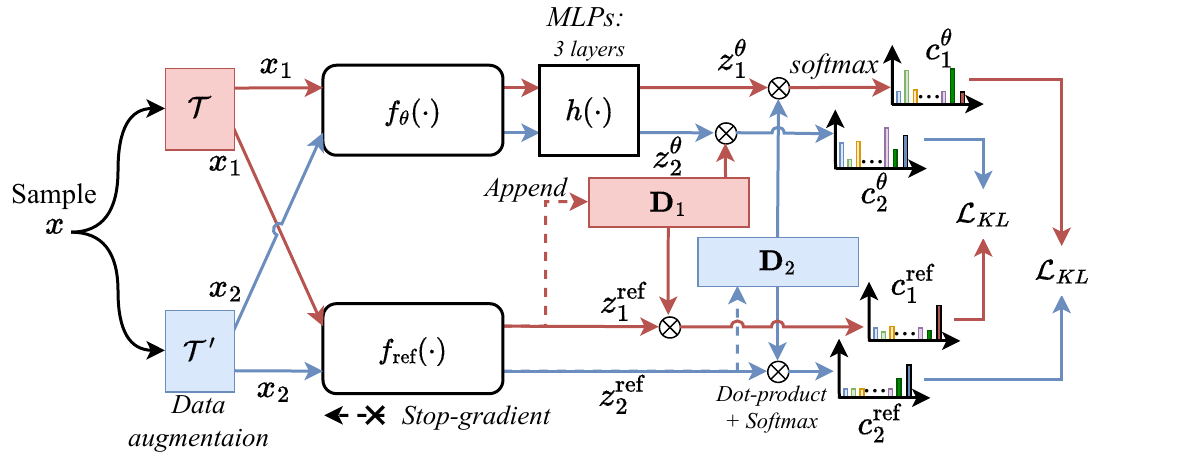}
\vspace{-9mm}
\caption{The overview of \emph{Self-Supervised Cross-View Training (SCT)}.}
\vspace{-4mm}
\label{fig:overall}
\end{figure*}

\subsection{Learning From Distribution}
A recent approach from computer vision to mitigating the false negative problem is replacing binary labels with a distribution of similarity scores. 
The main idea is to compare samples $a$ and $p$ using similarity scores computed from the same collection of instances $\mathbf{D}$ as soft labels.
In particular, the discrepancy between $a$ and $p$ is expressed as the similarity score discrepancy. 
\citet{DBLP:conf/iclr/FangWWZYL21} proposed a knowledge distillation method by training a student network to imitate the similarity score distribution formulated by a teacher network.
\citet{tejankar-etal-2021-isd} introduced a distribution learning paradigm using a similarity distribution score inferred by a momentum encoder over a set of instances.
\citet{zheng-etat-2021-ressl} proposed a representation learning technique by modeling the relationship distribution between weak and contrastive augmentation schemes. 
These works' experimental results demonstrated higher performance than contrastive learning and avoided false negatives. 

\subsection{Summary}

As discussed in Section~\ref{subsec:contrast}, the main drawback of contrastive learning is the binary distinction between positive and negative samples.
Near duplicates can be mistakenly used as negative samples. 
While there exist learning methods that use only positive samples, they are still outperformed by contrastive learning.

Based on various experimental studies, the paradigm of learning from distribution shows promising results compared to the other two approaches. 
However, we have found that the direct application of distribution learning to our problem does \emph{not} yield consistent performance improvement (see Table~\ref{tab:ablation_stuides} in Section~\ref{subsub:model_loss}).
We developed a self-referencing mechanism through data augmentation, which is needed to improve the distribution learning strategy for self-supervised sentence representation learning.  
Moreover, our framework also allows sentence representation learning in a distillation manner.
In particular, we employ a larger model as the teacher model to let a smaller model mimics the teacher's property.

\section{Proposed Method} \label{section:propose_method}
One of the challenges of using small models is the limited number of parameters.  
An empirical study has shown that larger models have enough parameters to solve complex problems with simple techniques, while smaller ones require more guidance to solve complex problems~\cite{pmlr-v97-brutzkus19b,DBLP:conf/nips/WangW0B0020}. 
Based on this observation, we design our proposed solution, \emph{Self-Supervised Cross-View Training (SCT)}, to enhance the learning guidance for smaller models (those with less than 100M parameters) by improving the self-referencing and discrepancy measurement mechanisms.

Figure~\ref{fig:overall} illustrates the SCT pipeline and highlights the two mechanisms we introduce to improve the learning guidance: \emph{cross-view comparison pipeline} and \emph{similarity-score-distribution Learning}.
In what follows, we describe how the cross-view pipeline improves the robustness of the self-referencing mechanism in Section~\ref{sec:crossview}.
Section~\ref{sec:simscoredist} presents the mechanism we use to measure the discrepancies between cross-view outputs.
We explain our proposed SCT loss function in Section~\ref{sec:loss}.
Finally, we introduce sentence representation distillation into our proposed framework in Section~\ref{subsec:distillation}.

\subsection{Cross-View Comparison Pipeline}
\label{sec:crossview}

As stated in the introduction, we devise a cross-view comparison pipeline to improve the robustness of the self-referencing mechanism.
Figure~\ref{fig:overall} illustrates how the two augmented views are fed to both online (updatable) $f_\theta(\cdot)$ and reference (un-updatable) $f_\text{ref}(\cdot)$ networks and how their outputs are compared in a cross-view pattern.
In this way, we use both views as references and do not compare outputs originating from the same view to each other.  

Given a new sample $x$, two augmentations $\mathcal{T}$ and $\mathcal{T}'$ are created from two different back-translations to produce two views $x_1=\mathcal{T}(x)$ and $x_2=\mathcal{T}'(x)$. 
Our framework allows various data augmentation schemes, i.e., masked language model (MLM) or Synonym replacement.  
We found that back-translation improves the performance of downstream tasks the most, and we used them to create cross-view inputs. (see Section~\ref{subsubsec:data_augment} for design analysis).

\noindent
\textbf{Online representations ($z^{\theta}$)}. The views $x_1$ and $x_2$ are first encoded by an encoder $f_\theta(\cdot)$ into a sentence representation, which is then mapped by the MLPs projector $h(\cdot)$ onto the representations $z^{\theta}_{1} = h(f_\theta(x_1))$ and $z^{\theta}_{2} = h(f_\theta(x_2))$.

\noindent
\textbf{Reference representations ($z^{\text{ref}}$)}.
The views $x_1$ and $x_2$ are again encoded by the $f_\text{ref}(\cdot)$ encoder to be used as references for the next step $z^{\text{ref}}_{1} = f_\text{ref}(x_1)$ and $z^{\text{ref}}_{2} = f_\text{ref}(x_2)$.
Note that the architecture and weights of the target network $f_\text{ref}(\cdot)$ and the online network $f_\theta(\cdot)$ are identical, and all encoder outputs are normalized.

\noindent
\textbf{Instance queues ($\mathbf{D}$)}. We denote two instance queues that are formulated from the cross-view reference representations, $z^{\text{ref}}_{1}$ and $z^{\text{ref}}_{2}$, as $\mathbf{D}_{1} = [\mathbf{d}^{\text{1}}_{1},...,\mathbf{d}^{k}_{1}]$ and $\mathbf{D}_{2} = [\mathbf{d}^{1}_{2},...,\mathbf{d}^{k}_{2}]$ where $k$ is the queue length and $\mathbf{d}$ is the sentence vector obtained from $f_\text{ref}$ with $\mathbf{d}^{k}_{1} = z^{\text{ref}}_{1}$ and $\mathbf{d}^{k}_{2} = z^{\text{ref}}_{2}$.
These instance queues enable the dynamic construction of a large and consistent negative sample, facilitating distribution learning.~\cite{DBLP:conf/iclr/FangWWZYL21,tejankar-etal-2021-isd,zheng-etat-2021-ressl}. 
At the beginning of each minibatch, we enqueue and dequeue instance queues in a ``first-in-first-out" manner.

\subsection{Similarity-Score Distribution}
\label{sec:simscoredist}
The next step is to calculate similarity score distributions for cross-view comparison. 
As shown in Figure~\ref{fig:overall}, we enforce the online representations $z^{\theta}_{1}$ and $z^{\theta}_{2}$ to maintain the consistency of the reference representations $z^{\text{ref}}_{2}$ and $z^{\text{ref}}_{1}$ through instance queues $\mathbf{D}_{2}$ and $\mathbf{D}_{1}$, respectively.
When the online network can match the reference representation in a large number of negative samples, the online network gains robustness to unseen inputs, which is necessary for sentence embedding.

We formulate the cross-view and reference distributions as follows:
\begin{compactitem}[\hspace{\setalign}•]
    \item We formulate a cross-view distribution that compares two augmented views called $c^{\theta}_{1} = \mathbf{SR}(z^{\theta}_{1}, \mathbf{D}_{2}, \tau^{\theta})$.
    \item Similarly, we derive another cross-view distribution called $c^{\theta}_{2} = \mathbf{SR}(z^{\theta}_{2}, \mathbf{D}_{1}, \tau^{\theta})$.
    \item We calculate the self-references with respect to the previous online distributions as follows: $c^{\text{ref}}_{1} = \mathbf{SR}(z^{\text{ref}}_{1}, \mathbf{D}_{1}, \tau^{\text{ref}})$ and $c^{\text{ref}}_{2} = \mathbf{SR}(z^{\text{ref}}_{2}, \mathbf{D}_{2}, \tau^{\text{ref}})$.
\end{compactitem}
We define the similarity score distribution function $\mathbf{SR}(\cdot)$ as a dot product function between a sentence representation and an instance queue:
%
\begin{equation} 
\begin{aligned}
\mathbf{SR}\left(z, \mathbf{D}, \tau \right)= \left[p_{1} \ldots p_{k}\right], \\
\quad \text{ where } p_{j}=\frac{e^{\text{sim}(z,\mathbf{d}_{j}) / \tau}} {\sum_{\mathbf{d} \sim \mathbf{D}} e^{\text{sim}(z,\mathbf{d}) / \tau}},
\end{aligned}
\label{eq:softmax}
\end{equation}
%
and $\tau$ is the temperature scaling hyper-parameter separately for the \emph{online} and \emph{reference} representations, and $\text{sim}(\cdot)$ is the dot product similarity function.

\subsection{Self-Supervised Cross-View Training Loss}
\label{sec:loss}
This step computes the self-supervised cross-view training $\mathcal{L}_{\text{SCT}}$ loss function using cross-view and reference distributions.
In particular, the loss is computed by minimizing the discrepancy between the $c^{\theta}_{1}$ and $c^{\text{ref}}_{2}$ distributions.
Moreover, we minimize the difference between the $c^{\theta}_{2}$ and $c^{\text{ref}}_{1}$ distributions.
$\mathcal{L}_{\text{SCT}}$ is defined as follows:
%
\begin{equation} 
\begin{aligned}
\mathcal{L}_{\text{SCT}} = \frac{1}{2} \mathcal{L}_{KL}(\text{SG}(c^{\text{ref}}_{2})||c^{\theta}_{1}) \\
+ \frac{1}{2} \mathcal{L}_{KL}(\text{SG}(c^{\text{ref}}_{1})||c^{\theta}_{2}),
\end{aligned}
\label{eq:final_loss_v2}
\end{equation}
%
given that $\mathcal{L}_{KL}$ is the KL-divergence loss function that minimizes the discrepancy between online and reference cross-view distributions.
Using stop-gradient $\text{SG}(\cdot)$ on the reference encoder is essential in avoiding the anisotropic problem (every input generates the same output).
As demonstrated in previous sentence embedding works~\cite{li-etal-2020-sentence,yan-etal-2021-consert}, when directly adapting BERT to STS tasks, the model tends to produce high similarity scores for all sentences, as it maps all sentences into a small region of the embedding space, also known as a ``collapse''. 
Many works have offered explanations for why the stop-gradient can help prevent the collapse issue in self-supervised training. \citet{DBLP:conf/iclr/ZhangZZPYK22} demonstrated the advantages of stop-gradient is in enhancing the gradient during the training process, resulting in de-centering and de-correlation effects. In addition, \citet{Chen_2021_CVPR} and \citet{DBLP:conf/cvpr/TaoWZDSHD22} also reported conforming results. 

With the $\mathcal{L}_{\text{SCT}}$'s mechanism, \emph{cross-view comparison pipeline} and \emph{similarity-score-distribution learning}, we circumvent the anisotropic problem that occurred in regular contrastive learning methods.

\subsection{Representation Distillation} \label{subsec:distillation}
As discussed in Section~\ref{subsec:sent_distil}, distillation is a common technique for improving the performance of the small PLMs by minimizing the discrepancy between the teacher model (large PLM) and the student model (small PLM).
In this work, we incorporate the distillation approach into our novel cross-view framework by replacing the reference network $f_\text{ref}(\cdot)$ with a larger PLM $f_\text{large}(\cdot)$ to enable the framework to perform the distillation.
We then design the distillation training objective by combining a self-supervised loss ($\mathcal{L}_{\text{SCT}}$) and a cross-view distillation loss ($\mathcal{L}_{\text{CD}}$) as follows:
\begin{equation} 
\begin{aligned}
\mathcal{L} =  \underbrace{\mathcal{L}_{\text{SCT}}}_{\text{self-supervised}}+\underbrace{\mathcal{L}_{\text{CD}}}_{\text{distillation}},
\end{aligned}
\label{eq:final_loss}
\end{equation}
given the $\mathcal{L}_{\text{SCT}}$ loss is a self-supervised consistency training loss based on the self-referencing mechanism, the  $\mathcal{L}_{\text{CD}}$ loss is a minimization objective between the large $f_\text{large}(\cdot)$ and small $f_{\theta}(\cdot)$ PLMs using the cross-view training pipeline. 
This loss aims to ensure that the small PLM can generate sentence representations similar to the large PLM.
We define $\mathcal{L}_{\text{CD}}$ as: 
\begin{equation} 
\begin{aligned}
\mathcal{L}_{\text{CD}} = \frac{1}{2} \mathcal{L}_{KL}(\text{SG}(c^{\text{large}}_{2})||c^{\theta}_{1}) \\
+ \frac{1}{2} \mathcal{L}_{KL}(\text{SG}(c^{\text{large}}_{1})||c^{\theta}_{2}),
\end{aligned}
\label{eq:final_loss_v3}
\end{equation}
We formulate $c^{\theta}$ from the small PLM (Section~\ref{sec:simscoredist}). 
We define $c^{\text{large}}_{1}=\mathbf{SR}(z^{\text{large}}_{1}, \mathbf{D}_{1}, \tau^{\text{ref}})$ and $c^{\text{large}}_{2}=\mathbf{SR}(z^{\text{large}}_{2}, \mathbf{D}_{2}, \tau^{\text{ref}})$.
We produce $z^{\text{large}}$ from $f_\text{large}(\cdot)$, where the input of $f_\text{large}(\cdot)$ is the same $x_1$ and $x_2$ from Section~\ref{sec:crossview}.
In addition, we apply the stop-gradient technique to prevent the large network from mimicking the online network.

\section{Experimental Settings}
\subsection{Implementation Details}

\textbf{Architecture.}
Our experiments cover five BERT PLMs~\cite{turc2019,DBLP:conf/naacl/DevlinCLT19}, while the number of parameters is ranged between 4M and 340M.
%
%
To obtain sentence representation vectors, we follow the practice of average word pooling presented by \citet{reimers-gurevych-2019-sentence}.
The projection head $h(\cdot)$ contains three MLP layers.
Each MLP layer has one feed-forward with a ReLU activation function, which is then fed into a linear feed-forward layer.
The size of the first and second feed-forward layers are $uw$ and $u$, respectively, where $u$ is the output vector dimension and $w$ is the first-second layer expansion factor.
The default value of $w$ is set to 10.

%
%
%

\noindent
\textbf{Training setup.}
For the training data, we use unlabeled texts from two NLI datasets, such as SNLI~\cite{bowman-etal-2015-large} and MultiNLI~\cite{williams-etal-2018-broad} datasets, following the prior works~\cite{li-etal-2020-sentence, DBLP:conf/emnlp/ZhangHLLB20, zhang-etal-2021-bootstrapped}.
For augmentation schemes, we use English-German-English $\mathcal{T}$ and English-French-English $\mathcal{T}'$ back-translations from \citet{zhang-etal-2021-bootstrapped}.
We use AdamW~\cite{LoshchilovAdamW} as the optimizer, a linear learning rate warm-up over 10\% of the training data, and a batch size of 128 for ten epochs.
We tune the learning rate, instance queue's size $k$, and the temperature scaling $\tau^{\theta}$ and $\tau^{\text{ref}}$ on the STS-B development set.
The best values of these parameters are shown in Table~\ref{tab:best_parameter}.
Note that we evaluate the STS-B development set every 64 training steps, and the best checkpoint is used for the final model.
We also initialize the queues by randomly generating vectors.
%

%
\begin{table}[h]
\definecolor{colorsmall}{rgb}{1,1,1}
\definecolor{colormedium}{rgb}{1,1,1}
\definecolor{colorlarge}{rgb}{1,1,1}
\hspace*{-5mm}
\centering
\setlength\doublerulesep{3pt}
\scalebox{0.9}{
\setlength{\tabcolsep}{3.5pt}
\begin{tabular}{l r|c|c|c|c}
\hline
\multicolumn{2}{c|}{\textbf{Model (\#parameters)}} & \textbf{LR} & \textbf{$k$} & \textbf{$\tau^\text{ref}$} & \textbf{$\tau^\theta$} \\  \hline \hline
\smallmodel BERT-Tiny   & 
\smallmodel (4M)    & 5E-04 & 131072 & 0.03 & 0.04 \\ 
\smallmodel BERT-Mini   & 
\smallmodel  (11M)   & 3E-04 & 131072 & 0.01 & 0.03 \\ 
\mediummodel BERT-Small  & 
\mediummodel (29M)  & 3E-04 & 65536  & 0.02 & 0.03 \\ 
\largemodel BERT-Base   & 
\largemodel (110M)  &  5E-04 & 65536  & 0.04 & 0.05 \\ 
\largemodel BERT-Large  & 
\largemodel (340M) & 5E-04 & 16384  & 0.04 & 0.05 \\
  \hline \hline
\end{tabular}
}
\vspace{-2mm}
\caption{\label{font-table} Model parameters, including learning rate, instance queue size $k$, and temperature scaling for reference $\tau^\text{ref}$ and online $\tau^\theta$ networks.}
\vspace{-5mm}
\label{tab:best_parameter}
\end{table}

\subsection{Competitive Methods}
We compare our work with a comprehensive range of self-supervised sentence representation methods representing well-known approaches discussed in Section~\ref{section_related_work}. 
\begin{compactitem}[\hspace{\setalign}•]
    \item \textbf{SimCSE}~\cite{gao-etal-2021-simcse}. A contrastive learning technique using different random dropout masks in the transformer architecture as the data augmentation.
     \item \textbf{DCLR}~\cite{zhou-etal-2022-debiased}. A contrastive learning method that weights negative samples according to the difficulty given by another model.
    \item \textbf{DiffCSE}~\cite{chuang-etal-2022-diffcse}. A contrastive learning technique that uses additional learning signals from a discriminator to make the model more sensitive to small changes. For the generator model used in this baseline, we employ DistilBERT~\cite{Sanh-etal-2019-distilbert} as described in the original paper.
    \item \textbf{CKD}~\cite{wu2021disco}. A self-supervised contrastive distillation method using a memory bank as large-negative samples.
    \item \textbf{ConGen}~\cite{limkonchotiwat-etal-2022-congen}. A self-supervised distillation method using an instance queue for distilling sentence embedding from large to small PLMs.
\end{compactitem}

\subsection{Evaluation Setup}
We utilize \citet{gao-etal-2021-simcse}'s evaluation settings by evaluating the efficiency of our work on the following STS benchmark datasets: STS-B~\cite{cer-etal-2017-semeval}, SICK-R~\cite{marelli-etal-2014-sick}, and STS 2012-2016~\cite{agirre-etal-2012-semeval,agirre-etal-2013-sem,agirre-etal-2014-semeval,agirre-etal-2015-semeval,agirre-etal-2016-semeval}. 
These datasets contain pair-wise sentences, where the similarity of each pair is labeled with a number between 0 and 5, indicating the degree to which the two sentences express the same meaning.

We also evaluate our model on downstream tasks, such as re-ranking (AskUbuntu~\cite{lei-etal-2016-semi} and SciDocs~\cite{specter2020cohan}) and NLI (SICK-E~\cite{marelli-etal-2014-sick} and SNLI~\cite{bowman-etal-2015-large} datasets).
For re-ranking, we use the experiment and evaluation settings from unsupervised sentence embedding benchmark~\cite{wang-etal-2021-tsdae-using}.
For NLI, we use all the datasets from SentEval~\cite{DBLP:conf/lrec/ConneauK18} and use the experiment setting from previous sentence embedding works~\cite{DBLP:conf/lrec/ConneauK18,limkonchotiwat-etal-2022-congen}. 
In addition, we report the average scores across three random seeds for each experiment where the SD value is approximately only $\sim$0.30 points for the STS benchmark, $\sim$1.02 points for NLI, and $\sim$0.78 points for NLI.

\begin{table*}[h]
\definecolor{colorsmall}{rgb}{1,1,1}
\definecolor{colormedium}{rgb}{1,1,1}
\definecolor{colorlarge}{rgb}{1,1,1}

\hspace*{-4mm}
\centering
\setlength\doublerulesep{3pt}

\scalebox{0.78}{
\setlength{\tabcolsep}{3pt}
\begin{tabular}{c|c|l|ccccccccc}
\hline
\multirow{2}{*}{\textbf{\begin{tabular}[c]{@{}c@{}}Type\\ \end{tabular}}} &
  \multirow{2}{*}{\textbf{\begin{tabular}[c]{@{}c@{}}Model\\ (\#parameters)\end{tabular}}} &
  \multicolumn{1}{c|}{\multirow{2}{*}{\textbf{Methods}}} &
  \multicolumn{8}{c}{\textbf{Semantic Textual Similarity (STS)}} \\ \cline{4-12} 
 &
   &
  \multicolumn{1}{c|}{} &
  \multicolumn{1}{c|}{\textbf{STS12}} &
  \multicolumn{1}{c|}{\textbf{STS13}} &
  \multicolumn{1}{c|}{\textbf{STS14}} &
  \multicolumn{1}{c|}{\textbf{STS15}} &
  \multicolumn{1}{c|}{\textbf{STS16}} &
  \multicolumn{1}{c|}{\textbf{STS-B}} &
  \multicolumn{1}{c|}{\textbf{SICK-R}} &
  \multicolumn{1}{c}{\textbf{Avg.}} &
  \\ \hline \hline
\multirow{20}{*}{Fine-tuning} &
  \smallmodel &
  SimCSE &
  \multicolumn{1}{c|}{58.59} &
  \multicolumn{1}{c|}{69.52} &
  \multicolumn{1}{c|}{60.15} &
  \multicolumn{1}{c|}{69.93} &
  \multicolumn{1}{c|}{67.85} &
  \multicolumn{1}{c|}{61.77} &
  \multicolumn{1}{c|}{60.27} &
  64.47 \\ 
 &
  \smallmodel &
  DCLR &
  \multicolumn{1}{c|}{52.39} &
  \multicolumn{1}{c|}{60.91} &
  \multicolumn{1}{c|}{50.57} &
  \multicolumn{1}{c|}{59.61} &
  \multicolumn{1}{c|}{56.49} &
  \multicolumn{1}{c|}{47.31} &
  \multicolumn{1}{c|}{55.68} &
  54.71 \\ 
 &
 \smallmodel  &
  DiffCSE &
  \multicolumn{1}{c|}{59.40} &
  \multicolumn{1}{c|}{\textbf{71.28}} &
  \multicolumn{1}{c|}{61.21} &
  \multicolumn{1}{c|}{71.85} &
  \multicolumn{1}{c|}{67.65} &
  \multicolumn{1}{c|}{61.78} &
  \multicolumn{1}{c|}{59.70} &
  64.70 \\ 
 &
  \multirow{-4}{*}{\smallmodel \begin{tabular}[c]{@{}c@{}}  BERT-Tiny\\ (4 M)\end{tabular}} &
  SCT &
  \multicolumn{1}{c|}{\textbf{70.67}} &
  \multicolumn{1}{c|}{66.68} &
  \multicolumn{1}{c|}{\textbf{66.76}} &
  \multicolumn{1}{c|}{\textbf{77.66}} &
  \multicolumn{1}{c|}{\textbf{70.62}} &
  \multicolumn{1}{c|}{\textbf{71.79}} &
  \multicolumn{1}{c|}{\textbf{63.95}} &
  \textbf{69.73} \\ \cline{2-11} 
 &
  \smallmodel &
  SimCSE &
  \multicolumn{1}{c|}{56.55} &
  \multicolumn{1}{c|}{65.77} &
  \multicolumn{1}{c|}{59.55} &
  \multicolumn{1}{c|}{72.26} &
  \multicolumn{1}{c|}{70.23} &
  \multicolumn{1}{c|}{60.85} &
  \multicolumn{1}{c|}{62.19} &
  65.94 \\ 
 &
  \smallmodel &
  DCLR &
  \multicolumn{1}{c|}{46.43} &
  \multicolumn{1}{c|}{60.44} &
  \multicolumn{1}{c|}{53.03} &
  \multicolumn{1}{c|}{65.12} &
  \multicolumn{1}{c|}{62.67} &
  \multicolumn{1}{c|}{51.83} &
  \multicolumn{1}{c|}{58.81} &
  56.90 \\ 
 &
  \smallmodel &
  DiffCSE &
  \multicolumn{1}{c|}{58.84} &
  \multicolumn{1}{c|}{\textbf{68.61}} &
  \multicolumn{1}{c|}{62.12} &
  \multicolumn{1}{c|}{74.59} &
  \multicolumn{1}{c|}{72.34} &
  \multicolumn{1}{c|}{64.84} &
  \multicolumn{1}{c|}{62.74} &
  66.30 \\ 
 &
  \multirow{-4}{*}{\smallmodel \begin{tabular}[c]{@{}c@{}}BERT-Mini\\ (11M)\end{tabular}} &
  SCT &
  \multicolumn{1}{c|}{\textbf{69.68}} &
  \multicolumn{1}{c|}{66.90} &
  \multicolumn{1}{c|}{\textbf{65.35}} &
  \multicolumn{1}{c|}{\textbf{78.29}} &
  \multicolumn{1}{c|}{\textbf{72.48}} &
  \multicolumn{1}{c|}{\textbf{69.47}} &
  \multicolumn{1}{c|}{\textbf{64.98}} &
  \textbf{69.59} \\ \cline{2-11} 

 &
  \mediummodel &
  SimCSE &
  \multicolumn{1}{c|}{60.34} &
  \multicolumn{1}{c|}{\textbf{73.84}} &
  \multicolumn{1}{c|}{66.28} &
  \multicolumn{1}{c|}{76.31} &
  \multicolumn{1}{c|}{73.94} &
  \multicolumn{1}{c|}{69.04} &
  \multicolumn{1}{c|}{64.13} &
  69.13 \\ 
 &
  \mediummodel &
  DCLR &
  \multicolumn{1}{c|}{56.81} &
  \multicolumn{1}{c|}{70.57} &
  \multicolumn{1}{c|}{60.12} &
  \multicolumn{1}{c|}{70.90} &
  \multicolumn{1}{c|}{69.03} &
  \multicolumn{1}{c|}{61.67} &
  \multicolumn{1}{c|}{62.87} &
  64.57 \\ 
 &
  \mediummodel &
  DiffCSE &
  \multicolumn{1}{c|}{59.35} &
  \multicolumn{1}{c|}{ 70.95} &
  \multicolumn{1}{c|}{65.24} &
  \multicolumn{1}{c|}{76.57} &
  \multicolumn{1}{c|}{73.21} &
  \multicolumn{1}{c|}{67.86} &
  \multicolumn{1}{c|}{64.82} &
  68.29 \\ 
 &
  \multirow{-4}{*}{\mediummodel \begin{tabular}[c]{@{}c@{}}BERT-Small\\ (29M)\end{tabular}}  &
  SCT &
  \multicolumn{1}{c|}{\textbf{70.98}} &
  \multicolumn{1}{c|}{69.89} &
  \multicolumn{1}{c|}{\textbf{69.50}} &
  \multicolumn{1}{c|}{\textbf{81.43}} &
  \multicolumn{1}{c|}{\textbf{75.26}} &
  \multicolumn{1}{c|}{\textbf{75.33}} &
  \multicolumn{1}{c|}{\textbf{65.52}} &
  \textbf{72.56}  \\ \cline{2-11}
 &\largemodel
  & SimCSE &
  \multicolumn{1}{c|}{68.40} &
  \multicolumn{1}{c|}{82.41} &
  \multicolumn{1}{c|}{74.38} &
  \multicolumn{1}{c|}{80.91} &
  \multicolumn{1}{c|}{78.56} &
  \multicolumn{1}{c|}{76.85} &
  \multicolumn{1}{c|}{\textbf{72.23}} &
  76.25 \\ 
 &\largemodel
  & DCLR &
  \multicolumn{1}{c|}{70.81} &
  \multicolumn{1}{c|}{83.73} &
  \multicolumn{1}{c|}{75.11} &
  \multicolumn{1}{c|}{72.56} &
  \multicolumn{1}{c|}{78.44} &
  \multicolumn{1}{c|}{78.31} &
  \multicolumn{1}{c|}{71.59} &
  77.22 \\ 
 &\largemodel
  & DiffCSE &
  \multicolumn{1}{c|}{72.78} &
  \multicolumn{1}{c|}{\textbf{84.43}} &
  \multicolumn{1}{c|}{\textbf{76.47}} &
  \multicolumn{1}{c|}{\textbf{83.90}} &
  \multicolumn{1}{c|}{\textbf{80.54}} &
  \multicolumn{1}{c|}{\textbf{80.59}} &
  \multicolumn{1}{c|}{71.23} &
  \textbf {78.49} \\ 
 & \multirow{-4}{*}{\largemodel \begin{tabular}[c]{@{}c@{}}BERT-Base\\ (110M)\end{tabular}} &
  SCT &
  \multicolumn{1}{c|}{\textbf{78.83}} &
  \multicolumn{1}{c|}{78.02} &
  \multicolumn{1}{c|}{72.64} &
  \multicolumn{1}{c|}{82.42} &
  \multicolumn{1}{c|}{76.12} &
  \multicolumn{1}{c|}{76.91} &
  \multicolumn{1}{c|}{68.89} &
  75.55 \\ \cline{2-11}
 &\largemodel &
  SimCSE &
  \multicolumn{1}{c|}{70.88} &
  \multicolumn{1}{c|}{84.16} &
  \multicolumn{1}{c|}{76.43} &
  \multicolumn{1}{c|}{84.50} &
  \multicolumn{1}{c|}{79.76} &
  \multicolumn{1}{c|}{79.26} &
  \multicolumn{1}{c|}{73.88} &
  78.41 \\ 
 &\largemodel &
  DCLR &
  \multicolumn{1}{c|}{71.87} &
  \multicolumn{1}{c|}{\textbf{84.83}} &
  \multicolumn{1}{c|}{\textbf{77.37}} &
  \multicolumn{1}{c|}{84.70} &
  \multicolumn{1}{c|}{\textbf{79.81}} &
  \multicolumn{1}{c|}{\textbf{79.55}} &
  \multicolumn{1}{c|}{\textbf{74.19}} &
  \textbf{78.90} \\ 
 &\largemodel &
  \begin{tabular}[c]{@{}l@{}}DiffCSE\\ (reproduce)\end{tabular} &
  \multicolumn{1}{c|}{71.82} &
  \multicolumn{1}{c|}{84.39} &
  \multicolumn{1}{c|}{75.85} &
  \multicolumn{1}{c|}{\textbf{84.97}} &
  \multicolumn{1}{c|}{79.20} &
  \multicolumn{1}{c|}{\textbf{79.55}} &
  \multicolumn{1}{c|}{73.42} &
  78.46 \\ 
 & \multirow{-5}{*}{\largemodel \begin{tabular}[c]{@{}c@{}}BERT-Large\\ (340M)\end{tabular}} &
  SCT &
  \multicolumn{1}{c|}{\textbf{76.61}} &
  \multicolumn{1}{c|}{81.80} &
  \multicolumn{1}{c|}{76.84} &
  \multicolumn{1}{c|}{84.34} &
  \multicolumn{1}{c|}{77.15} &
  \multicolumn{1}{c|}{78.85} &
  \multicolumn{1}{c|}{71.55} &
  \multicolumn{1}{c}{78.16} \\ \hline  
  \hline

  \multirow{12}{*}{Distillation} &
  \smallmodel &
  CKD &
  \multicolumn{1}{c|}{71.76} &
  \multicolumn{1}{c|}{\textbf{80.41}} &
  \multicolumn{1}{c|}{73.63} &
  \multicolumn{1}{c|}{81.75} &
  \multicolumn{1}{c|}{76.14} &
  \multicolumn{1}{c|}{75.89} &
  \multicolumn{1}{c|}{67.78} &
  75.34 \\ 
 &
  \smallmodel &
  ConGen &
  \multicolumn{1}{c|}{71.76} &
  \multicolumn{1}{c|}{79.75} &
  \multicolumn{1}{c|}{73.47} &
  \multicolumn{1}{c|}{82.53} &
  \multicolumn{1}{c|}{76.64} &
  \multicolumn{1}{c|}{78.01} &
  \multicolumn{1}{c|}{\textbf{69.19}} &
  75.89 \\ 
 &
  \multirow{-3}{*}{\smallmodel \begin{tabular}[c]{@{}c@{}}  BERT-Tiny\\ (4 M)\end{tabular}} &
  SCT &
  \multicolumn{1}{c|}{\textbf{73.46}} &
  \multicolumn{1}{c|}{80.14} &
  \multicolumn{1}{c|}{\textbf{73.95}} &
  \multicolumn{1}{c|}{\textbf{82.97}} &
  \multicolumn{1}{c|}{\textbf{77.24}} &
  \multicolumn{1}{c|}{\textbf{78.44}} &
  \multicolumn{1}{c|}{68.92} &
  \textbf{76.43} \\ \cline{2-11} 
 &
  \smallmodel &
  CKD &
  \multicolumn{1}{c|}{72.39} &
  \multicolumn{1}{c|}{\textbf{81.98}} &
  \multicolumn{1}{c|}{75.37} &
  \multicolumn{1}{c|}{82.83} &
  \multicolumn{1}{c|}{77.71} &
  \multicolumn{1}{c|}{77.73} &
  \multicolumn{1}{c|}{67.58} &
  76.51 \\ 
 &
  \smallmodel &
  ConGen &
  \multicolumn{1}{c|}{72.96} &
  \multicolumn{1}{c|}{81.15} &
  \multicolumn{1}{c|}{74.46} &
  \multicolumn{1}{c|}{83.11} &
  \multicolumn{1}{c|}{77.07} &
  \multicolumn{1}{c|}{79.46} &
  \multicolumn{1}{c|}{69.48} &
  76.81 \\ 
 &
  \multirow{-3}{*}{\smallmodel \begin{tabular}[c]{@{}c@{}}BERT-Mini\\ (11M)\end{tabular}} &
  SCT &
  \multicolumn{1}{c|}{\textbf{74.49}} &
  \multicolumn{1}{c|}{81.14} &
  \multicolumn{1}{c|}{\textbf{75.53}} &
  \multicolumn{1}{c|}{\textbf{84.18}} &
  \multicolumn{1}{c|}{\textbf{77.83}} &
  \multicolumn{1}{c|}{\textbf{80.04}} &
  \multicolumn{1}{c|}{\textbf{69.84}} &
  \textbf{77.58} \\ \cline{2-11} 

 &
  \mediummodel &
  CKD &
  \multicolumn{1}{c|}{72.43} &
  \multicolumn{1}{c|}{82.11} &
  \multicolumn{1}{c|}{75.59} &
  \multicolumn{1}{c|}{82.19} &
  \multicolumn{1}{c|}{77.73} &
  \multicolumn{1}{c|}{77.21} &
  \multicolumn{1}{c|}{68.05} &
  76.47 \\ 
 &
  \mediummodel &
  ConGen &
  \multicolumn{1}{c|}{73.61} &
  \multicolumn{1}{c|}{82.37} &
  \multicolumn{1}{c|}{74.93} &
  \multicolumn{1}{c|}{83.19} &
  \multicolumn{1}{c|}{77.77} &
  \multicolumn{1}{c|}{79.54} &
  \multicolumn{1}{c|}{69.73} &
  77.31  \\ 
 &
  \multirow{-3}{*}{\mediummodel \begin{tabular}[c]{@{}c@{}}BERT-Small\\ (29M)\end{tabular}}  &
  SCT &
  \multicolumn{1}{c|}{\textbf{74.96}} &
  \multicolumn{1}{c|}{\textbf{82.83}} &
  \multicolumn{1}{c|}{\textbf{75.89}} &
  \multicolumn{1}{c|}{\textbf{84.08}} &
  \multicolumn{1}{c|}{\textbf{78.24}} &
  \multicolumn{1}{c|}{\textbf{80.53}} &
  \multicolumn{1}{c|}{\textbf{70.57}} &
  \textbf{78.16}  \\ \cline{2-11}
 &\largemodel
  & CKD &
  \multicolumn{1}{c|}{72.52} &
  \multicolumn{1}{c|}{84.37} &
  \multicolumn{1}{c|}{76.79} &
  \multicolumn{1}{c|}{82.97} &
  \multicolumn{1}{c|}{79.01} &
  \multicolumn{1}{c|}{78.21} &
  \multicolumn{1}{c|}{69.26} &
  77.58 \\ 
 &\largemodel
  & ConGen &
  \multicolumn{1}{c|}{74.15} &
  \multicolumn{1}{c|}{84.24} &
  \multicolumn{1}{c|}{76.72} &
  \multicolumn{1}{c|}{84.76} &
  \multicolumn{1}{c|}{79.11} &
  \multicolumn{1}{c|}{80.78} &
  \multicolumn{1}{c|}{71.31} &
  78.72 \\ 
 & \multirow{-3}{*}{\largemodel \begin{tabular}[c]{@{}c@{}}BERT-Base\\ (110M)\end{tabular}} &
  SCT &
  \multicolumn{1}{c|}{\textbf{76.60}} &
  \multicolumn{1}{c|}{\textbf{84.72}} &
  \multicolumn{1}{c|}{\textbf{77.63}} &
  \multicolumn{1}{c|}{\textbf{85.19}} &
  \multicolumn{1}{c|}{\textbf{79.68}} &
  \multicolumn{1}{c|}{\textbf{81.28}} &
  \multicolumn{1}{c|}{\textbf{71.97}} &
  \textbf{79.58} \\ \cline{2-11}
\hline  
  \hline 
\end{tabular}}
\vspace{-2mm}
\caption{\label{font-table} Sentence embedding performance on STS tasks (Spearman's rank correlation). For the distillation setting, we used BERT-Large-SimCSE for all distillation techniques.}
\vspace{-5mm}
\label{tab:sts_tasks}
\end{table*}
\section{Experimental Results}

This section presents results from five sets of studies.
Section~\ref{subsec:sts_small} presents results from the main experiments using 7 STS benchmark datasets described in the previous subsection. 
In Section~\ref{subsec:downstream_task}, we demonstrate the effectiveness of our method on various downstream benchmark datasets. 
In Section~\ref{subsec:design}, we study the design decisions of the key components, namely (i) the model architecture and loss function; (ii) instance queues; and (iii) data augmentation strategy. 
Section~\ref{subsec:distillation_study} demonstrates the design decision of the distillation loss.

\subsection{Main Results: STS Benchmark Datasets}
\label{subsec:sts_small}
Table~\ref{tab:sts_tasks} illustrates the effectiveness of our method \emph{(SCT)} in comparison to the five competitors: SimCSE, CCLR, DiffCSE, CKD, and ConGen.
We separate the results into two groups: fine-tuning (without a large PLM in the framework) and distillation (using a large PLM in the framework).

\noindent
\textbf{Fine-tuning results}. For the average scores, the experimental results show that our method SCT outperforms all competitors for all PLMs with less than 100M parameters. 
Let us first look at the results from BERT-Tiny, the smallest one from the BERT family.
SCT outperforms SimCSE and DCLR by 5.26 and 4.3 points regarding Spearman's rank correlation.
As expected, SCT is outperformed by competitors for models with more than 100M parameters, i.e., BERT-Base and BERT-Large.
For BERT-Base, SCT scores lower than the best performer, DiffCSE, by 2.94 points.
For BERT-Large, SCT scores lower than DCLR, which is the best performer, by 0.74 points.
These findings underscore the importance of incorporating SCT into PLM training, especially in scenarios where computational resources are limited.
%

\noindent
\textbf{Distillation results}. 
The results presented in Table~\ref{tab:sts_tasks} demonstrate that SCT outperforms competing methods across all PLMs. 
Notably, SCT shows superior performance compared to ConGen, with improvements from 75.89 to 76.43 and 78.72 to 79.58 on BERT-Tiny and BERT-Base, respectively. 
Furthermore, the SCT method outperforms the teacher model (BERT-Large-SimCSE) when the number of parameters exceeds 100M, achieving a Spearman's rank correlation of 79.58 compared to 78.41. 
This success can be attributed to the combination of the self-supervised loss $\mathcal{L}{\text{SCT}}$ and the distillation loss $\mathcal{L}{\text{CD}}$.
The efficacy of SCT is further demonstrated by comparing fine-tuning and distillation methods using the smallest PLM. 
The results show that the distillation method significantly improves the fine-tuning performance, with a boost from 69.73 to 76.43 (SCT fine-tuning and distillation).

\noindent
\textbf{Summary of results}. As shown in Table~\ref{tab:sts_tasks}, the fine-tuning experiments demonstrate that SCT outperforms its competitors 20 out of 35 times, representing 57.1\% of all cases.
For models with less than 100M parameters, SCT performs the best in 18 out of 21 trials, i.e., 85.7\%. 
In contrast, for models with more than 100M parameters, SCT is the top performer in only 2 out of 14 cases, i.e., 14.3\%.
In the distillation setting, SCT outperforms its competitors in 25 out of 28 experiments, i.e., 89.3\%, for all models.
Moreover,  when the number of parameters surpasses 29M, SCT is the best performer in all 14 cases. 
In addition, the performance of SCT-Distillation-BERT-Small (\#param: 29M) is similar to the SOTA on BERT-Base, i.e., 78.49 (DiffCSE) vs. 78.16 (SCT).
\emph{These results conform with the proposed benefit of SCT that we aim to improve the performance of smaller models.}

\subsection{Downstream tasks} \label{subsec:downstream_task}
In this study, we demonstrate the effectiveness of our method compared to DCLR and DiffCSE (the top performers in Table~\ref{tab:sts_tasks}) on re-ranking (AskUbuntu and SciDocs) and natural language inference (SICK-E and SNLI).
We report the Mean Average Precision (MAP) for re-ranking and accuracy score for NLI.
In addition, we also separate the results into two groups just like in the previous section.

\noindent
\textbf{Fine-tuning results}. Table~\ref{tab:downstream_tasks} demonstrates that while SCT's performance on STS is lower than that of its competitors when the parameter count is less than 100M, it outperforms all competitors in re-ranking and NLI for 26 out of 28 cases (92.8\%).  
For example, on BERT-large, SCT surpasses DiffCSE and DCLR by 2.52 and 2.64 points in the NLI average case, respectively.
The gap between our method and competitive methods is wider on NLI datasets compared to STS benchmark datasets. 
For re-ranking, we found that SCT consistently outperforms competitive methods except for AskUbuntu on BERT-Large.
These results demonstrate that SCT improves the robustness of any PLMs on downstream tasks with cross-view and self-referencing mechanisms. 

\noindent
\textbf{Distillation results}. The results indicate that SCT outperforms all competing distillation methods. 
Furthermore, our distillation method performs better than the fine-tuning method in comparable setups.
For instance, when applied to the smallest PLM (BERT-Tiny), our distillation method improved the performance of NLI datasets from 71.89 to 78.53, outperforming the fine-tuning method. 
Moreover, SCT-Distillation-BERT-Base surpasses SOTA BERT-Large-finetuning for the average case. 
These findings highlight the efficacy of SCT in improving PLM performance, whether the teacher model is available or not.

\begin{table}[ht]
\hspace*{-4mm}
\centering
\setlength\doublerulesep{1.5pt}
\scalebox{0.7}{
\setlength{\tabcolsep}{2pt}
\begin{tabular}{c|c|l|ccc|ccc}
\hline
\multirow{2}{*}{\textbf{Type}} &
  \multirow{2}{*}{\textbf{\begin{tabular}[c]{@{}c@{}}Model\\ (BERT)\end{tabular}}}  &
  \multicolumn{1}{c|}{\multirow{2}{*}{\textbf{Methods}}} &
  \multicolumn{3}{c|}{\textbf{Reranking}} &
  \multicolumn{3}{c}{\textbf{NLI}} \\ \cline{4-9} 
 &
   &
  \multicolumn{1}{c|}{} &
  \multicolumn{1}{c|}{\textbf{AskU}} &
  \multicolumn{1}{c|}{\textbf{SciDocs}} &
  \textbf{Avg} &
  \multicolumn{1}{c|}{\textbf{SICK-E}} &
  \multicolumn{1}{c|}{\textbf{SNLI}} &
  \textbf{Avg.} \\ \hline \hline
\multirow{12}{*}{\rotatebox[origin=c]{90}{Fine-tuning}} &
  \multirow{3}{*}{Tiny} &
  DCLR &
  \multicolumn{1}{c|}{50.23} &
  \multicolumn{1}{c|}{58.52} &
  54.38 &
  \multicolumn{1}{c|}{74.93} &
  \multicolumn{1}{c|}{63.09} &
  69.01 \\ 
 &
   &
  DiffCSE &
  \multicolumn{1}{c|}{50.76} &
  \multicolumn{1}{c|}{59.26} &
  55.01 &
  \multicolumn{1}{c|}{74.45} &
  \multicolumn{1}{c|}{63.36} &
  68.91 \\ 
 &
   &
  SCT &
  \multicolumn{1}{c|}{\textbf{51.26}} &
  \multicolumn{1}{c|}{\textbf{59.32}} &
  \textbf{55.29} &
  \multicolumn{1}{c|}{\textbf{78.18}} &
  \multicolumn{1}{c|}{\textbf{65.60}} &
  \textbf{71.89} \\ \cline{2-9} 
 &
  \multirow{3}{*}{Small} &
  DCLR &
  \multicolumn{1}{c|}{51.09} &
  \multicolumn{1}{c|}{62.89} &
  56.99 &
  \multicolumn{1}{c|}{80.24} &
  \multicolumn{1}{c|}{69.17} &
  74.71 \\ 
 &
   &
  DiffCSE &
  \multicolumn{1}{c|}{51.56} &
  \multicolumn{1}{c|}{63.02} &
  57.29 &
  \multicolumn{1}{c|}{80.29} &
  \multicolumn{1}{c|}{69.02} &
  74.66 \\ 
 &
   &
  SCT &
  \multicolumn{1}{c|}{\textbf{51.76}} &
  \multicolumn{1}{c|}{\textbf{65.41}} &
  \textbf{58.59} &
  \multicolumn{1}{c|}{\textbf{80.37}} &
  \multicolumn{1}{c|}{\textbf{71.02}} &
  \textbf{75.70} \\ \cline{2-9} 
 &
  \multirow{3}{*}{Base} &
  DCLR &
  \multicolumn{1}{c|}{51.29} &
  \multicolumn{1}{c|}{69.47} &
  60.38 &
  \multicolumn{1}{c|}{81.19} &
  \multicolumn{1}{c|}{71.60} &
  76.40 \\ 
 &
   &
  DiffCSE &
  \multicolumn{1}{c|}{50.93} &
  \multicolumn{1}{c|}{69.33} &
  60.13 &
  \multicolumn{1}{c|}{\textbf{82.30}} &
  \multicolumn{1}{c|}{72.57} &
  77.44 \\ 
 &
   &
  SCT &
  \multicolumn{1}{c|}{\textbf{52.40}} &
  \multicolumn{1}{c|}{\textbf{69.54}} &
  \textbf{60.97} &
  \multicolumn{1}{c|}{\textbf{82.30}} &
  \multicolumn{1}{c|}{\textbf{73.56}} &
  \textbf{77.93} \\ \cline{2-9} 
 &
  \multirow{3}{*}{Large} &
  DCLR &
  \multicolumn{1}{c|}{\textbf{53.79}} &
  \multicolumn{1}{c|}{72.36} &
  \textbf{63.08} &
  \multicolumn{1}{c|}{81.83} &
  \multicolumn{1}{c|}{71.99} &
  76.91 \\ 
 &
   &
  DiffCSE &
  \multicolumn{1}{c|}{52.60} &
  \multicolumn{1}{c|}{68.78} &
  60.69 &
  \multicolumn{1}{c|}{81.77} &
  \multicolumn{1}{c|}{72.29} &
  77.03 \\  
 &
   &
  SCT &
  \multicolumn{1}{c|}{53.35} &
  \multicolumn{1}{c|}{\textbf{72.68}} &
  63.02 &
  \multicolumn{1}{c|}{\textbf{83.54}} &
  \multicolumn{1}{c|}{\textbf{75.56}} &
  \textbf{79.55} \\ \hline
\multirow{9}{*}{\rotatebox[origin=c]{90}{Distillation}} &
  \multirow{3}{*}{Tiny} &
  CKD &
  \multicolumn{1}{c|}{55.40} &
  \multicolumn{1}{c|}{65.04} &
  60.22 &
  \multicolumn{1}{c|}{82.46} &
  \multicolumn{1}{c|}{72.36} &
  77.41 \\ 
 &
   &
  ConGen &
  \multicolumn{1}{c|}{56.27} &
  \multicolumn{1}{c|}{64.75} &
  60.51 &
  \multicolumn{1}{c|}{83.07} &
  \multicolumn{1}{c|}{73.07} &
  78.07 \\ 
 &
   &
  SCT &
  \multicolumn{1}{c|}{\textbf{56.76}} &
  \multicolumn{1}{c|}{\textbf{65.52}} &
  \textbf{61.14} &
  \multicolumn{1}{c|}{\textbf{83.16}} &
  \multicolumn{1}{c|}{\textbf{73.90}} &
  \textbf{78.53} \\ \cline{2-9} 
 &
  \multirow{3}{*}{Small} &
  CKD &
  \multicolumn{1}{c|}{56.04} &
  \multicolumn{1}{c|}{67.73} &
  61.89 &
  \multicolumn{1}{c|}{82.87} &
  \multicolumn{1}{c|}{74.27} &
  78.57 \\ 
 &
   &
  ConGen &
  \multicolumn{1}{c|}{54.99} &
  \multicolumn{1}{c|}{67.93} &
  61.46 &
  \multicolumn{1}{c|}{83.17} &
  \multicolumn{1}{c|}{75.77} &
  79.47 \\ 
 &
   &
  SCT &
  \multicolumn{1}{c|}{\textbf{55.65}} &
  \multicolumn{1}{c|}{\textbf{68.22}} &
  \textbf{61.94} &
  \multicolumn{1}{c|}{\textbf{84.11}} &
  \multicolumn{1}{c|}{\textbf{76.76}} &
  \textbf{80.44} \\ \cline{2-9} 
 &
  \multirow{3}{*}{Base} &
  CKD &
  \multicolumn{1}{c|}{56.85} &
  \multicolumn{1}{c|}{70.53} &
  63.69 &
  \multicolumn{1}{c|}{83.13} &
  \multicolumn{1}{c|}{74.04} &
  79.05 \\ 
 &
   &
  ConGen &
  \multicolumn{1}{c|}{56.70} &
  \multicolumn{1}{c|}{71.25} &
  63.98 &
  \multicolumn{1}{c|}{83.48} &
  \multicolumn{1}{c|}{76.28} &
  79.88 \\ 
 &
   &
  SCT &
  \multicolumn{1}{c|}{\textbf{57.40}} &
  \multicolumn{1}{c|}{\textbf{71.85}} &
  \textbf{64.63} &
  \multicolumn{1}{c|}{\textbf{84.73}} &
  \multicolumn{1}{c|}{\textbf{77.82}} &
  \textbf{80.97} \\ \hline \hline
\end{tabular}
}
\vspace{-2mm}
\caption{\label{font-table} Re-ranking and NLI results. We report MAP scores for re-ranking and accuracy for NLI.}
\vspace{-5mm}
\label{tab:downstream_tasks}
\end{table}

\subsection{Design Analysis}
\label{subsec:design}
In this subsection, we analyze the key components of SCT as follows.
Section~\ref{subsub:model_loss} provides an ablation study on the model and loss function.
Section~\ref{subsub:instance_queue} presents an analysis of the instance queue.
Section~\ref{subsubsec:data_augment} explores how different data augmentation schemes affect the performance of our method. 
In section~\ref{subsubsec:summary_ablation}, we provide the summary of results from the design analysis studies. 
\begin{table}[h]
\vspace{-1mm}
\hspace*{-4mm}
\centering
\setlength\doublerulesep{3pt}
\scalebox{0.8}{
\setlength{\tabcolsep}{2pt}
\begin{tabular}{lcccc}
\hline
\multicolumn{1}{c|}{\textbf{Method}} &
  \multicolumn{1}{c|}{\smallmodel{\textbf{BERT-Tiny}}} &
  \multicolumn{1}{c}{\mediummodel{\textbf{BERT-Small}}} &\\ \hline \hline
\multicolumn{1}{l|}{SCT} &
  \multicolumn{1}{c|}{\textbf{69.73}} &
  \multicolumn{1}{c}{\textbf{72.56}} & \\ \hline
\multicolumn{3}{l}{\textbf{\textit{Model \& loss studies}}} \\ \hline
\multicolumn{1}{l|}{\begin{tabular}[c]{@{}l@{}}Distribution $\rightarrow$ Contrastive\\\end{tabular}} &
  \multicolumn{1}{c|}{\begin{tabular}[c]{@{}c@{}}{\color[HTML]{FE0000}$\downarrow$5.30}\\ \end{tabular}} &
  \multicolumn{1}{c}{\begin{tabular}[c]{@{}c@{}}{\color[HTML]{FE0000}$\downarrow$10.52}\\ \end{tabular}} &
  \\
\multicolumn{1}{l|}{Cross-view$\rightarrow$Identical-view} &
  \multicolumn{1}{c|}{\begin{tabular}[c]{@{}c@{}}{\color[HTML]{FE0000}$\downarrow$7.82}\\ \end{tabular}} &
  \multicolumn{1}{c}{\begin{tabular}[c]{@{}c@{}}{\color[HTML]{FE0000}$\downarrow$3.69}\\ \end{tabular}} &
\\
\multicolumn{1}{l|}{\begin{tabular}[c]{@{}l@{}}$f_\text{ref} \rightarrow$ a momentum encoder\\ \end{tabular}} &
  \multicolumn{1}{c|}{{\color[HTML]{FE0000}$\downarrow$1.29}} &
  \multicolumn{1}{c}{{\color[HTML]{FE0000}$\downarrow$1.06}} &
    \\ 
\multicolumn{1}{l|}{KL $\rightarrow$ CE} &
  \multicolumn{1}{c|}{{\color[HTML]{FE0000}$\downarrow$0.20}} &
  \multicolumn{1}{c}{{\color[HTML]{FE0000}$\downarrow$0.32}} &
  \\ 
\multicolumn{1}{l|}{No MLPs} &
  \multicolumn{1}{c|}{{\color[HTML]{FE0000}$\downarrow$1.77}} &
  
  \multicolumn{1}{c}{{\color[HTML]{FE0000}$\downarrow$3.67}} &
    \\  \hline
\multicolumn{3}{l}{\textbf{\textit{Instance queue studies}}} \\ \hline
\multicolumn{1}{l|}{Only one instance queue} &
  \multicolumn{1}{c|}{{\color[HTML]{FE0000}$\downarrow$2.85}} &
  \multicolumn{1}{c}{{\color[HTML]{FE0000}$\downarrow$1.21}} &
  \\ 
\multicolumn{1}{l|}{No update queues} &
  \multicolumn{1}{c|}{\begin{tabular}[c]{@{}c@{}}{\color[HTML]{FE0000}$\downarrow$3.85}\\ \end{tabular}} &
  \multicolumn{1}{c}{\begin{tabular}[c]{@{}c@{}}{\color[HTML]{FE0000}$\downarrow$2.76}\\ \end{tabular}} &
  \\ 
\multicolumn{1}{l|}{\begin{tabular}[c]{@{}l@{}}No instance queue\\ \end{tabular}} &
  \multicolumn{1}{c|}{{\color[HTML]{FE0000}$\downarrow$3.38}} &
  \multicolumn{1}{c}{{\color[HTML]{FE0000}$\downarrow$2.29}} &
   \\ \hline
  
\multicolumn{3}{l}{\textbf{\textit{Data augmentation studies}}} \\ \hline
\multicolumn{1}{l|}{Masked language model} &
  \multicolumn{1}{c|}{{\color[HTML]{FE0000}$\downarrow$3.82}} &
  \multicolumn{1}{c}{{\color[HTML]{FE0000}$\downarrow$1.87}} &
    \\ 
\multicolumn{1}{l|}{Synonym replacement} &
  \multicolumn{1}{c|}{{\color[HTML]{FE0000}$\downarrow$7.95}} &
  \multicolumn{1}{c}{{\color[HTML]{FE0000}$\downarrow$3.15}} &
   \\ 
\multicolumn{1}{l|}{Dropout mask} &
  \multicolumn{1}{c|}{{\color[HTML]{FE0000}$\downarrow$5.42}} &
  
  \multicolumn{1}{c}{{\color[HTML]{FE0000}$\downarrow$3.11}} &
   \\ 
\multicolumn{1}{l|}{Using the same BT ($\mathcal{T}=\mathcal{T}'$)} &
  \multicolumn{1}{c|}{{\color[HTML]{FE0000}$\downarrow$5.66}} &
 
  \multicolumn{1}{c}{{\color[HTML]{FE0000}$\downarrow$2.47}} &
  \\ \hline \hline

\end{tabular}}
\vspace{-2mm}
\caption{\label{font-table} Ablation studies on model \& loss, instance queue studies, and data augmentation studies. We evaluate the performance of these studies on the average score across seven STS datasets.}
\vspace{-4mm}
\label{tab:ablation_stuides}
\end{table}

\subsubsection{Model and loss Function} 
\label{subsub:model_loss}
Table~\ref{tab:ablation_stuides} presents the results from the proposed SCT (fine-tuning) setup compared to the following variants.
For brevity, we focus on models with less than 100M parameters.

The results show that the default version of SCT is the best performer. 
We can see that changing from distribution learning to contrastive learning incurs performance penalties ranging from 3.24 to 10.52 points.
Similarly, changing the view comparison setting from cross-view to identical-view also results in performance penalties ranging from 3.69 to 13.06 points. 
In contrast, the momentum encoder, cross-entropy, and removing MLPs modifications result in smaller impacts.
The results suggest that all design components are crucial to our method's performance, and the penalties for removing the distribution learning and cross-view parts are the most drastic ones.

\subsubsection{Instance Queue}
\label{subsub:instance_queue}

We study the impact of the following instance queue modifications:
(i) combining two instance queues into one,
(ii) keeping the negative samples unchanged,
(iii) replacing the queues with in-batch negatives. 
As shown in Table~\ref{tab:ablation_stuides}, any modification from the default SCT results in a performance drop for all models.
We can also see that keeping the negative samples unchanged suffers the worst impact. 
For instance, the performance of using the same negative sample (no queue updates) decreases the performance from 69.73 to 65.88 on BERT-Tiny.
These results imply that the coverage of negative samples is crucial to the performance.

\begin{figure}[h]
\hspace*{-3.5mm}
\centering
\includegraphics[width=0.4\textwidth]{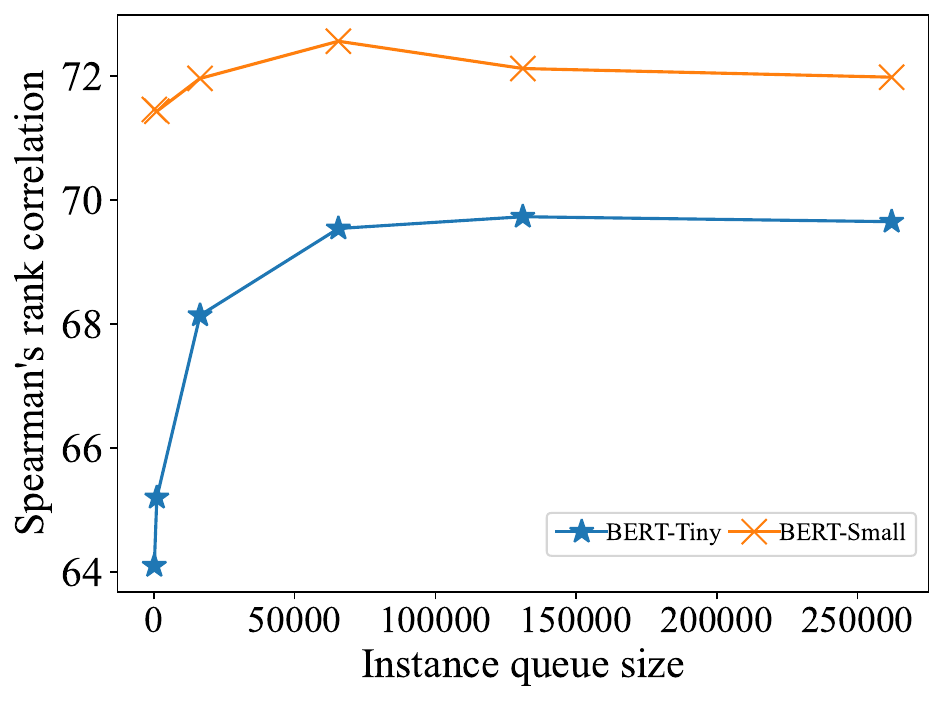}
\vspace{-2mm}
\caption{The comparison between various queue sizes, such as 128, 1024, 16384, 65536, 131072, and 262144. We average Spearman's rank correlation across the seven STS benchmarks and test on small PLMs, i.e., BERT-Tiny and BERT-Small.}
\vspace{-4mm}
\label{fig:queue_size_ablation}
\end{figure}

Let us now consider the impact of instance queue size on BERT-Tiny and BERT-Small.
In this study, we vary the number of negative samples in the queue from 128 to 262,144 samples (the largest that our hardware supports).
As shown in Figure~\ref{fig:queue_size_ablation}, the performance improves as the queue size grows from 128 to 16,384 samples for all cases.
However, the optimal queue size varies according to the model architecture, i.e., 131,072 for BERT-Tiny and 65,536 for BERT-Small.
These results suggest we should tune the queue size separately for each model architecture. 

\subsubsection{Data augmentation choice}
\label{subsubsec:data_augment}
This experiment evaluates the effect of different augmentation schemes widely used in sentence representation learning: (i) two back-translations (default), (ii) mask language model 15\%, (iii) synonym replacement (one-word replacement),  (iii) dropout mask, and (iv) using the same back-translation ($\mathcal{T}=\mathcal{T}'$).  
We evaluate Spearman's rank correlation on seven STS benchmark datasets.
The experimental results are shown in Table~\ref{tab:ablation_stuides} (Data augmentation studies).

As expected, changing back-translation to other augmentation schemes decreases the performance in all cases. 
For instance, the performance of BERT-Tiny drops from 69.73 to 64.07 when we change from two back-translations to only one back-translation.
This is because the two back-translation schemes generate high-quality synonym text pairs (different syntax but same meaning), which help sentence representation to distinguish positive and negative samples in the embedding space. 
In contrast, other augmentation techniques produce either incorrect or similar pair texts, which are not useful for sentence representation learning.

\noindent
\emph{Data augmentation analysis}. To validate our data augmentation strategy, we assess the syntax and semantic scores on our augmented datasets. 
We utilized the edit distance metric to evaluate the syntax changes (dissimilarity) in the augmented datasets compared to the original dataset. 
Additionally, we employed cosine similarity to evaluate the semantic consistency between the original and augmented embeddings. 
Our base encoders in this analysis were BERT-Tiny-SCT and BERT-Tiny-DiffCSE.

The results, as presented in Table~\ref{tab:augment_analysis}, revealed that although MLM produced the highest string dissimilarity, it failed to preserve semantic from the original texts, resulting in significant changes to syntax and semantic. 
In contrast, the synonym augment scheme exhibited higher embedding similarity than MLM, as it maintained the original texts to a greater extent, resulting in minimal changes to syntax and semantic.
Interestingly, back-translation produced favorable results in both string and embedding similarity. 
While the syntax was altered, the semantic remained unchanged, indicating reasonable performance in maintaining the core semantic meaning. 
While the string dissimilarity of back-translation was slightly lower than that of MLM (with only one character difference on average), back-translation achieved higher similarities in the base encoders' embeddings.

\begin{table}[h]
\vspace{-1mm}
\hspace*{-4mm}
\centering
\setlength\doublerulesep{3pt}
\scalebox{0.8}{
\setlength{\tabcolsep}{2pt}
\begin{tabular}{l|c|cc}
\hline
\multicolumn{1}{c|}{\multirow{2}{*}{\textbf{Method}}} & \multirow{2}{*}{\textbf{String Dissim.}} & \multicolumn{2}{c}{\textbf{Embedding Sim.}} \\ \cline{3-4} 
\multicolumn{1}{c|}{} &                & \multicolumn{1}{c|}{\textbf{SCT}}  & \textbf{DiffCSE} \\ \hline
Synonym                & 6.59 $\pm$ 4.30            & \multicolumn{1}{c|}{0.94}          & \textbf{0.97}    \\ 
MLM                    & \textbf{13.37 $\pm$ 7.67} & \multicolumn{1}{c|}{0.85}          & 0.94             \\ 
BT                     & 12.32 $\pm$  9.31         & \multicolumn{1}{c|}{\textbf{0.98}} & \textbf{0.97}    \\ \hline \hline
\end{tabular}}
\vspace{-2mm}
\caption{\label{font-table} We evaluate the string dissimilarity and embedding similarity on our training and augmentation datasets. For the \emph{string dissimilarity}, we use edit distance to evaluate the changes in the augmentation dataset. For the \emph{embedding similarity}, we use cosine similarity to evaluate the identical of the original and augmentation dataset.}
\vspace{-2mm}
\label{tab:augment_analysis}
\end{table}

These findings corroborate the results of our data augmentation choices, as shown in Table~\ref{tab:ablation_stuides}, where we emphasize that data augmentation methods with desirable properties exhibit high string dissimilarity and embedding similarity. 
The efficacy of back-translation, in particular, highlights its potential as a suitable data augmentation technique for preserving both syntax and semantic consistency, making it a promising technique for enhancing the performance of embedding space.

\subsubsection{Summary of Design Analysis.} \label{subsubsec:summary_ablation}
As shown in Table~\ref{tab:ablation_stuides}, we present the desired components in the SCT framework. 
We found that applying a technique from computer vision requires careful consideration of the architecture and data augmentation schemes.
The experimental results from the model and loss studies demonstrate that using contrastive learning similar to SimCSE~\cite{gao-etal-2021-simcse} or using a momentum encoder similar to MoCo~\cite{moco-paper,chen2020improved} produce poorer performance than our setting (small PLMs).
This is because of the fact that small PLMs necessitate more guidance, as discussed in Section~\ref{section:propose_method}.
Thus, the similarity-score-distribution learning paradigm employed in our framework demonstrates promising results in enhancing the performance of small PLMs.
However, it is worth noting that applying the similarity-score-distribution learning paradigm from \citet{DBLP:conf/iclr/FangWWZYL21} without making any adjustments adversely affects the model's performance more than any other setting, i.e., the performance of BERT-Tiny decreased by 7.82 points when we changed from cross-view (our work) to identical-view (computer vision).

Regarding the data augmentation studies (Section~\ref{subsubsec:data_augment}), we found that using two-back translations produced the most effective augmented sentences compared to MLM or synonym replacement.
With these findings, we require to adjust architectures, loss, and data augmentation from previous works, which achieved SOTA performance in small PLMs.
These insightful findings necessitate the adaptation of architectures, loss functions, and data augmentation approaches from prior works. 
By carefully considering these adjustments, we can further enhance the capabilities and efficiency of small PLMs in various NLP tasks.


\subsection{Distillation Studies} \label{subsec:distillation_study}
In this subsection, we study the components of our distillation method as follows.
In Section~\ref{subsub:distillation_design}, we provide an ablation study on the model and loss function.
Section~\ref{subsec:distillation_loss} presents an analysis of the distillation loss.

\subsubsection{Distillation Design} \label{subsub:distillation_design}
This study illustrates the efficacy of SCT within distillation settings. 
An ablation study has been meticulously conducted to elucidate that all constituent elements of SCT contribute to the overall performance. 
In particular, we demonstrate the ablation study of the self-supervised loss $\mathcal{L}_{\text{SCT}}$ in distillation settings using the setup from Table~\ref{tab:ablation_stuides}.

The findings in Table~\ref{tab:loss_ablation} highlight the importance of adhering to the default SCT configuration, as any departure from it incurs a notable performance decrement. 
The analysis distinctly reveals that the most substantial penalties arise from the alterations involving Distribution$\rightarrow$Contrastive and Cross-view$\rightarrow$Identical-view adjustments. 
These results emphasize all components of SCT contribute to performance improvement. 
Any deviation from the default SCT setting results in a performance penalty.

\begin{table}[h]
\centering
\scalebox{0.8}{
\begin{tabular}{lcc}
\hline
\multicolumn{1}{c|}{\textbf{Method}}                                   & \multicolumn{1}{c|}{\textbf{BERT-Tiny}} & \textbf{BERT-Small} \\ \hline \hline
\multicolumn{1}{l|}{SCT-distillation}                        & \multicolumn{1}{c|}{76.43} & 78.16 \\ \hline
\multicolumn{3}{l}{\textit{\textbf{Model \& loss studies}}}                                      \\ \hline
\multicolumn{1}{l|}{Distribution $\rightarrow$ Contrastive}  & \multicolumn{1}{c|}{{\color[HTML]{FE0000}$\downarrow$6.03}}  & {\color[HTML]{FE0000}$\downarrow$3.26}  \\ 
\multicolumn{1}{l|}{Cross-view $\rightarrow$ Identical-view} & \multicolumn{1}{c|}{{\color[HTML]{FE0000}$\downarrow$3.97}}  & {\color[HTML]{FE0000}$\downarrow$1.43}  \\ 
\multicolumn{1}{l|}{$f_{\text{ref}}$ $\rightarrow$ a momentum encoder} & \multicolumn{1}{c|}{{\color[HTML]{FE0000}$\downarrow$3.62}}               & {\color[HTML]{FE0000}$\downarrow$0.97}                \\ 
\multicolumn{1}{l|}{KL $\rightarrow$ CE}                     & \multicolumn{1}{c|}{{\color[HTML]{FE0000}$\downarrow$3.43}}  & {\color[HTML]{FE0000}$\downarrow$0.98}  \\ 
\multicolumn{1}{l|}{No MLPs}                                 & \multicolumn{1}{c|}{{\color[HTML]{FE0000}$\downarrow$4.14}}  & {\color[HTML]{FE0000}$\downarrow$2.46}  \\ \hline \hline
\end{tabular}}
\caption{Ablation studies on model \& loss of our distillation method. We evaluate the performance of these studies on the average score across seven STS datasets.}
\label{tab:loss_ablation}
\end{table}

\subsubsection{Distillation Loss} \label{subsec:distillation_loss}
This experiment demonstrates the efficacy of our novel approach involving self-supervised and distillation losses.
We investigate the impact of using a distillation loss alone and the benefits of integrating both distillation and self-supervised losses.
In particular, we explore the utility of our SCT loss as a bootstrapping mechanism for existing distillation methods. 
We also demonstrate a common distillation loss by minimizing the discrepancy between $z^{\text{large}}$ and $z^{\theta}$ with Mean Square Error ($\mathcal{L}_\text{MSE}$).

Table~\ref{tab:distillation_stuides} presents the experimental results for two scenarios: (i) using only a distillation loss and (ii) incorporating both self-supervised and distillation losses.  
In addition, we highlight the improvement with the up arrow ($\uparrow$).
Our experimental findings consistently demonstrate that including the SCT loss significantly enhances the performance of existing distillation methods across the board. 
For example, the SCT loss yields substantial performance boosts of 3.28 and 4.50 for $\mathcal{L}_{\text{CD}}$ and $\mathcal{L}_{\text{MSE}}$ methods on BERT-Tiny, respectively.
Moreover, we improve the performance of ConGen and CKD methods to a level comparable with $\mathcal{L}_{\text{CD}}+\mathcal{L}_{\text{SCT}}$.
We do this using SCT as the bootstrapping loss.
These results underscore the advantages of combining distillation and self-supervised losses to achieve enhanced performance in small PLMs.
Furthermore, our SCT loss demonstrates its efficacy as a reliable bootstrapping loss for distillation methods, highlighting its potential as a valuable tool for improving the performance of distillation-based approaches.

\begin{table}[h]
\vspace{-1mm}
\hspace*{-4mm}
\centering
\setlength\doublerulesep{3pt}
\scalebox{0.85}{
\setlength{\tabcolsep}{2pt}
\begin{tabular}{l|cl|cl}
\hline
\multicolumn{1}{c|}{\textbf{Method}}               & \multicolumn{2}{c|}{\textbf{BERT-Tiny}}     & \multicolumn{2}{c}{\textbf{BERT-Small}}    \\ \hline \hline
$\mathcal{L}_{\text{CD}}$                           & \multicolumn{1}{c}{71.93} &                & \multicolumn{1}{c}{74.41} &                \\ 
$\mathcal{L}_{\text{CD}}+\mathcal{L}_{\text{SCT}}$  & \multicolumn{1}{c}{76.43} & \color[HTML]{008000}{$\uparrow$4.50} & \multicolumn{1}{c}{78.16} & \color[HTML]{008000}{$\uparrow$3.75} \\ \hline \hline
$\mathcal{L}_{\text{MSE}}$                          & \multicolumn{1}{c}{71.42} &                & \multicolumn{1}{c}{73.16} &                \\ 
$\mathcal{L}_{\text{MSE}}+\mathcal{L}_{\text{SCT}}$ & \multicolumn{1}{c}{74.70} & \color[HTML]{008000}{$\uparrow$3.28} & \multicolumn{1}{c}{76.11} & \color[HTML]{008000}{$\uparrow$2.95} \\ \hline \hline
$\mathcal{L}_{\text{ConGen}}$                       & \multicolumn{1}{c}{75.89} &                & \multicolumn{1}{c}{77.31} &                \\ 
$\mathcal{L}_{\text{ConGen}}+\mathcal{L}_{\text{SCT}}$ & \multicolumn{1}{c}{76.36} & \color[HTML]{008000}{$\uparrow$0.47} & \multicolumn{1}{c}{77.87} & \color[HTML]{008000}{$\uparrow$0.56} \\ \hline \hline
$\mathcal{L}_{\text{CKD}}$                          & \multicolumn{1}{c}{75.34} &                & \multicolumn{1}{c}{76.47} &                \\ 
$\mathcal{L}_{\text{CKD}}+\mathcal{L}_{\text{SCT}}$ & \multicolumn{1}{c}{76.06} & \color[HTML]{008000}{$\uparrow$0.72} & \multicolumn{1}{c}{76.94} & \color[HTML]{008000}{$\uparrow$0.46} \\ \hline \hline
\end{tabular}}
\vspace{-2mm}
\caption{\label{font-table} We evaluate the performance of these studies on the average score across seven STS datasets.}
\vspace{-5mm}
\label{tab:distillation_stuides}
\end{table}

\section{Conclusion}
We propose a self-supervised sentence representation learning method called \emph{Self-Supervised Cross-View Training (SCT)}. 
The observation inspires our work that smaller models, when constructed in a self-supervised setting, tend to perform poorly or collapse altogether. 
We hypothesize that this problem can be addressed by providing additional learning guidance to facilitate the self-referencing mechanism in the self-supervised learning pipeline. 

Our work consists of three key contributions.
\emph{First}, at the framework level, 
we formulate a \emph{cross-view comparison} pipeline to improve the self-referencing mechanism by enabling cross-comparison between two input views. 
In addition, our framework allows using two input views formulated from the same or different PLMs.
\emph{Second}, to facilitate the learning process, we also design a new technique to measure the discrepancy between two cross-view outputs. 
Instead of comparing them directly, we use similarity score distributions.
\emph{Third}, we conducted extensive sets of experimental studies to compare our method against existing competitors and to analyze our design decisions.

The experimental results on the STS tasks show that our method dominates all competitors in the cases of PLMs with less than 100M parameters.
With the help of the distillation loss, our method improves the performance of small PLMs better than that of large PLMs. 
Moreover, our method outperforms competitive methods for all PLMs on the downstream tasks.
Furthermore, the results also confirm that the cross-view comparison pipeline and similarity score distribution comparison are crucial to performance improvement. 
These findings imply that smaller PLMs benefit from our judiciously designed guidance in a self-supervised setting.

\bibliography{tacl2021}
\bibliographystyle{acl_natbib}

\clearpage

\end{document}